%% file: ral_final.tex
\documentclass[journal]{IEEEtran}
\IEEEoverridecommandlockouts

\let\labelindent\relax
\input{configs}

\title{\LARGE \bf{}Understanding Physical Effects for Effective Tool-use}

\author{Zeyu Zhang\quad{}Ziyuan Jiao\quad{}Weiqi Wang\quad{}Yixin Zhu\quad{}Song-Chun Zhu\quad{}Hangxin Liu
\thanks{Manuscript received February 24, 2022; accepted: June 26, 2022. 
Date of publication July xx, 2022; date of current version July xx, 2022. 
This letter was recommended for publication by Editor Hanna Kurniawati upon evaluation of the Associate Editor and Reviewers' comments. (\textit{Zeyu Zhang and Ziyuan Jiao contributed equally to this work}.) (\textit{Corresponding author: Hangxin Liu}.)}
\thanks{Zeyu Zhang, Ziyuan Jiao, and Song-Chun Zhu are with the Beijing Institute for General Artificial Intelligence, (BIGAI), Beijing 100080, China, and with the Center for Vision, Cognition, Learning, and Autonomy (VCLA), Statistics Department, UCLA (e-mail: zeyuzhang@ucla.edu; zyjiao@ucla.edu; sczhu@stat.ucla.edu).}
\thanks{Weiqi Wang is with the Center for Vision, Cognition, Learning, and Autonomy (VCLA), Statistics Department, UCLA (e-mail: weiqi.wang@ucla.edu).}
\thanks{Yixin Zhu is with the Institute for Artificial Intelligence, Peking University, Beijing 100871, China (e-mail: yixin.zhu@pku.edu.cn).}
\thanks{Song-Chun Zhu is also with the Institute for Artificial Intelligence, Peking University, Beijing 100871, China.}
\thanks{Hangxin Liu is with the Beijing Institute for General Artificial Intelligence, (BIGAI), Beijing 100080, China (e-mail: liuhx@bigai.ai).}
\thanks{This letter has supplementary downloadable material available at https://doi.org/XXXX/XXXXXX, provided by the authors.}
\thanks{Digital Object Identifier (DOI): see top of this page.}}

\begin{document}

\maketitle

\begin{abstract}
We present a robot learning and planning framework that produces an effective tool-use strategy with the least joint efforts, capable of handling objects different from training. Leveraging a \ac{fem}-based simulator that reproduces fine-grained, continuous visual and physical effects given observed tool-use events, the essential physical properties contributing to the effects are identified through the proposed \ac{idsr} algorithm. We further devise an optimal control-based motion planning scheme to integrate robot- and tool-specific kinematics and dynamics to produce an effective trajectory that enacts the learned properties. In simulation, we demonstrate that the proposed framework can produce more effective tool-use strategies, drastically different from the observed ones in two exemplar tasks.
\end{abstract}%

\setstretch{0.96}
\section{Introduction}

\IEEEPARstart{A} robot extends its capability to a broader range of tasks by using tools. Unlike treating a tool as a part of the end-effector that commonly appears in industrial settings~\cite{asada1988direct,holladay2019force}, researchers have proposed various learning-based approaches that empower more adept tool-use behaviors. However, existing learning objectives either focus on low-level motions~\cite{kawaharazuka2021adaptive,saito2021select} without an explicit understanding of the tasks or on higher-level concepts with simplified motion patterns~\cite{akizuki2018tactile,qin2020keto,turpin2021gift}. As a result, robots are still far from producing situational tool-use strategies: Given a set of objects (typical tools or canonical objects), which one would be the best to accomplish the task? Once an object is chosen as the tool, how to efficiently use it given robot- and tool-specific kinematics and dynamics?

\begin{figure}[t!]
    \centering
    \includegraphics[width=\linewidth]{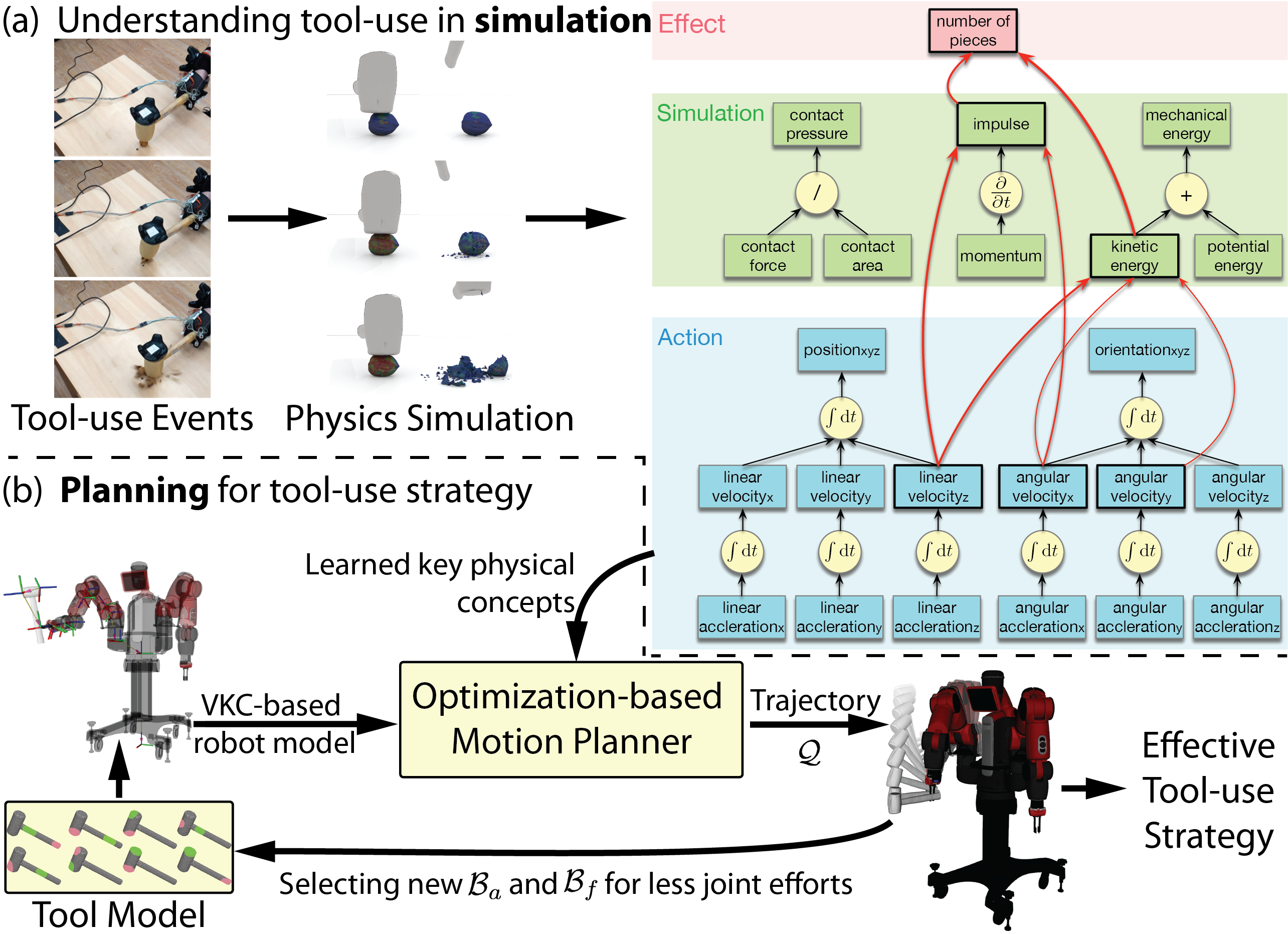}%
    \caption{\textbf{Overview of the proposed framework.} (a) After observing tool-use events, we learn the essential physical properties involved in the processes from the effects reproduced by physics-based simulation. (b) The learned results are formulated into a motion planning scheme to produce various strategies to use an object, and the most effective strategy with minimal joint efforts among others is selected.}
    \vspace{-9pt}
    \label{fig:motivation}
\end{figure}

To tackle these challenges, we propose an integrated learning and planning framework wherein robots understand and produce effective tool-use strategies by reasoning about the essential physical properties that contribute to the success of the task. \cref{fig:motivation} shows an overview of our integrated framework. Compared to prior arts in robotics literature, our framework identifies the invariant learning objective of tool-uses at a more fundamental level; instead of using pure vision-based methods~\cite{liu2017jointly,tang2020inferring}, our framework focuses on the physical effects produced by the tool and learns to recognize the essential physical properties in accomplishing the task. Specifically, we adopt a state-of-the-art \acf{fem}~\cite{li2020incremental} to simulate how both visual and physical effects evolve over time (\eg, stress, energy, contact) in a continuous manner. A symbolic regression-based \acf{idsr} algorithm is devised to trace the set of physical properties produced by the simulator and to efficiently identify how much each property contributes to the effect. 

Next, we formulate the learned results into an optimal control-based motion planning scheme that allows the robot to generate various tool-use strategies whose efficiency is evaluated by joint efforts. To ease the motion planning problem and make the scheme more generic (\ie, handle robots with different morphology, tools in diverse shapes, and various ways to operate tools), we introduce a \ac{vkc} perspective~\cite{jiao2021consolidated,jiao2021efficient} that treats the tool as an additional link of the robot and integrates their kinematic and dynamic properties as a whole in motion planning.

In two exemplar tasks\textemdash{}cracking walnut and cutting carrot, we demonstrate that the proposed learning and planning framework can (i) identify the essential physical properties significant to the success of the task and (ii) produce an effective tool-use strategy that emulates the essential properties while minimizing joint efforts using seen and unseen objects as tools. As a result, the proposed framework allows the robot to better understand the physical environment by leveraging physics-based simulations and become more competent in bootstrapping novel (\ie, not observed) tool-use strategies.

\setstretch{0.968}

\subsection{Related Work}

Learning tool-use involves several cognitive and intelligent processes, challenging even for humans. Replicating such a skill set at the full spectrum is thus difficult, and existing literature mainly focuses at one of three different levels. Low-level \textbf{planning and control} methods track desired tool-use trajectories with impedance control~\cite{asada1988direct}, alter force and motion constraints at different stages~\cite{holladay2019force}, or apply learning-based control~\cite{kawaharazuka2021adaptive,saito2021select}; robust execution is of the central interest. At mid-level, various \textbf{intermediate representations} are identified for better understanding tool-uses, such as keypoints~\cite{qin2020keto,turpin2021gift}, primitive parts~\cite{nair2019tool,nair2019autonomous,wicaksono2015learning,wicaksono2020cognitive}, and kinematic models~\cite{takahashi2017tool,jiao2021efficient}. Although introducing these representations facilitates learning more diverse tool-use skills, they are still restricted to the geometric association between shapes and task specifications. To capture high-level \textbf{concepts} embedded in tool-uses, researchers adopt task and motion planning~\cite{toussaint2018differentiable}, functionality and affordance~\cite{zhu2015understanding,akizuki2018tactile, levihn2015using}, causality~\cite{brawer2020causal}, and commonsense~\cite{allen2020rapid,tuli2021tango}, achieving better generalization capabilities. Empowered by physics-based simulation, we advance this line of work by taking all three views into account: (i) learning related physical properties as the concepts from the tasks at the high-level, (ii) integrating tool's properties to robots by adopting \ac{vkc} as the intermediate representation at the mid-level, and (iii) planning tool-use strategies via optimal control at the low-level.

Recently, physics-based \textbf{simulation} significantly facilitates various robotics tasks, \eg, Liu \etal simulate forces to bridge human and robot's embodiments~\cite{liu2019mirroring}, Kennedy \etal plan liquid pouring~\cite{kennedy2019autonomous}, Matl \etal infer granular materials' properties~\cite{matl2020inferring}, Hahn \etal approximate soft objects' motions by estimating visco-elastic parameters~\cite{hahn2019real2sim}, Geilinger \etal develop simulation framework for rigid and soft bodies with fictional contact to promote robot locomotion~\cite{geilinger2020add}, Li \etal improve UAV designs~\cite{li2021soft}, and Heiden \etal optimize robot's cutting and slicing motions~\cite{heiden2021disect}. Though sharing a similar spirit, the \ac{fem} simulator adopted in the paper~\cite{li2020incremental} is designed to produce a wider range of physical properties for robot learning instead of optimizing for dedicated applications.

\subsection{Overview}

The remainder of this paper is organized as follows. \cref{sec:definition} formally presents the problem definition and our optimal planning framework. \cref{sec:sim_learn} describes the simulation setup and the learning of physical properties. We demonstrate our framework's efficacy in \cref{sec:exp} and conclude the paper with discussions in \cref{sec:conclusion}.

\begin{figure}[t!]
    \centering
    \begin{subfigure}[b]{\linewidth}
        \centering  
        \includegraphics[width=\linewidth]{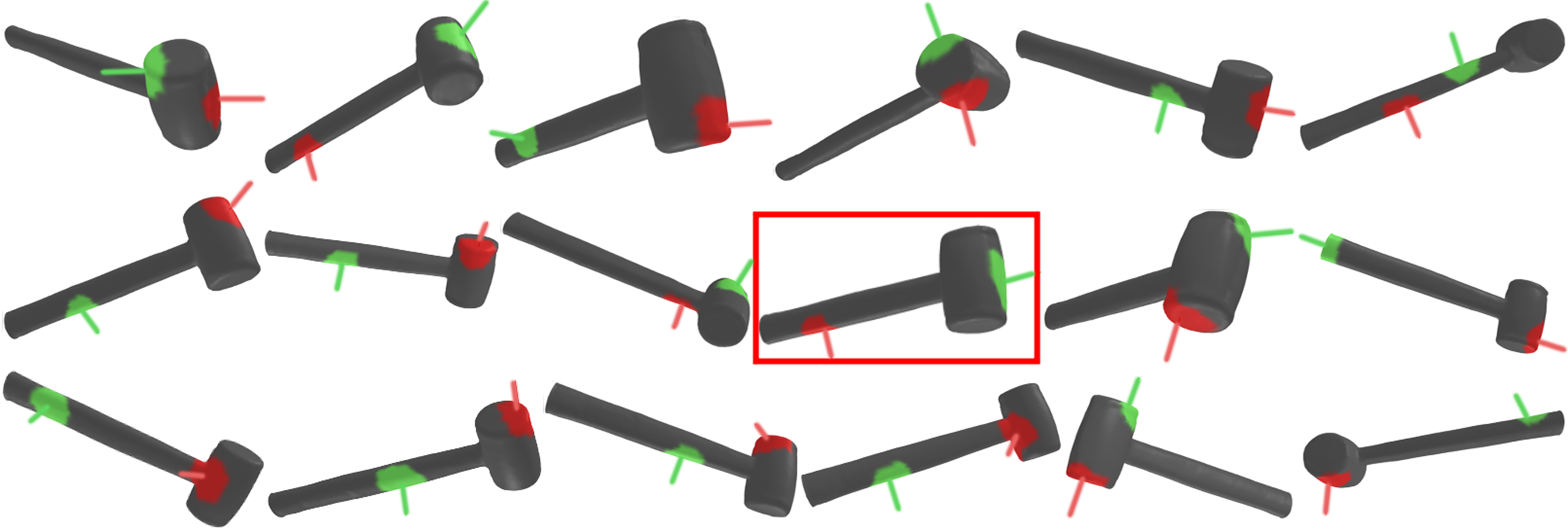}
        \caption{Various partitions of a hammer. The green regions denote affordance bases $\mathcal{B}_a$ as potential areas for grasps, whereas the red regions denote functional bases $\mathcal{B}_f$ as a candidate area to act on the target object. Surface normals calculated at the regions' center are the directions to grasp or act.}
        \label{fig:tool_basis}
    \end{subfigure}%
    \\
    \begin{subfigure}[b]{\linewidth}
        \centering  
        \includegraphics[width=\linewidth]{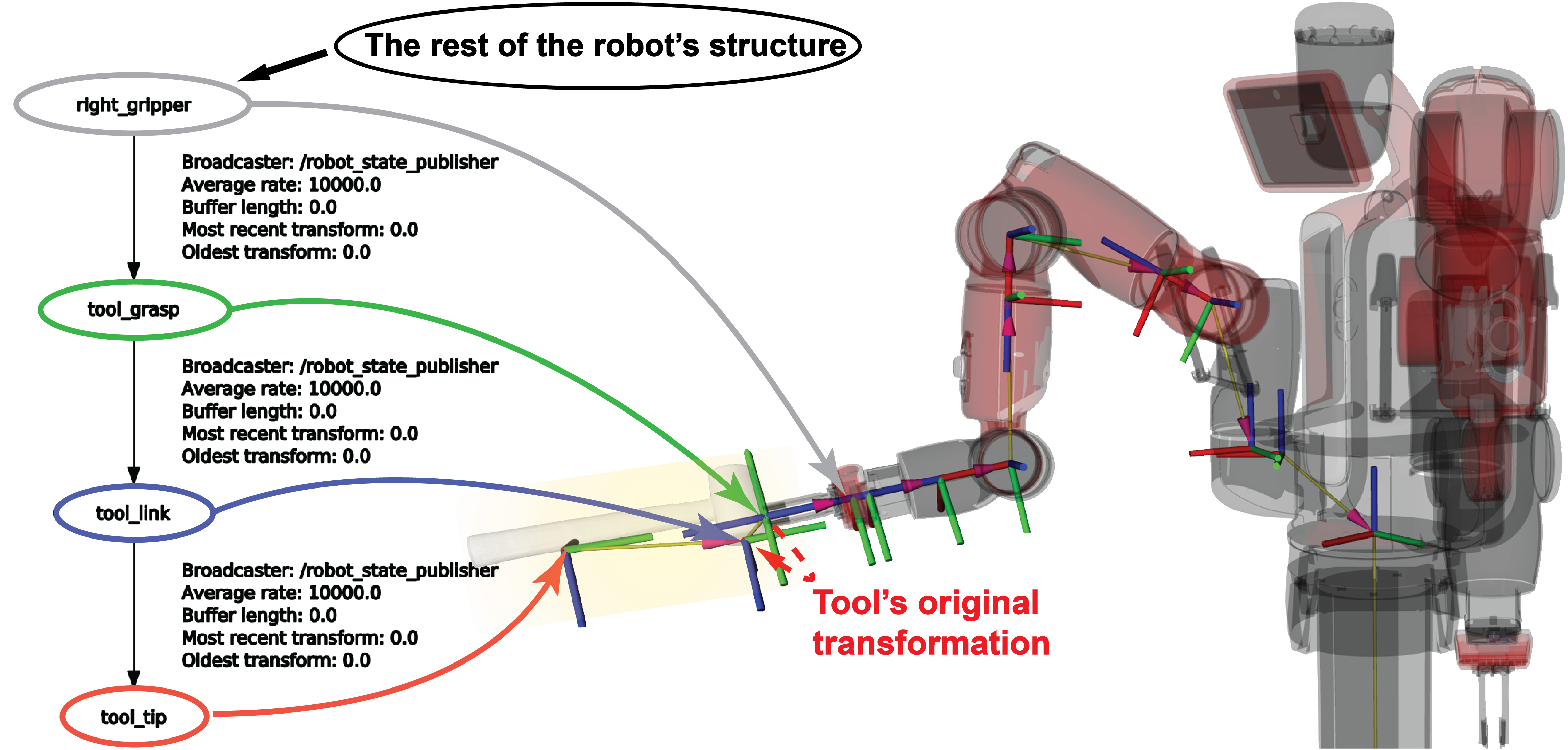}
        \caption{\Acp{vkc} can be constructed by consolidating the kinematic models with dynamic information of the robot and the tool.}
        \label{fig:robot_tool_vkc}
    \end{subfigure}%
    \caption{\textbf{A \Ac{vkc} perspective that promotes motion planning.} (a) Given a sampled bases combination (highlighted in red box) of $\mathcal{B}_a$ and $\mathcal{B}_f$, (b) a \ac{vkc} is constructed by assigning a virtual joint between the robot's gripper and the $\mathcal{B}_a$, and $\mathcal{B}_f$ becomes the new end-effector. This \ac{vkc} conversion and construction supports efficient and optimal motion planning to produces proper tool-use trajectories by taking both kinematic and dynamic factors into account.}
    \label{fig:vkc}
\end{figure}

\section{Problem Definition}\label{sec:definition}

We define a tool-use strategy $\mathcal{S} = (\mathcal{B}_a, \mathcal{B}_f, \mathcal{Q})$ by (i) an affordance basis $\mathcal{B}_a$ to be grasped by the robot gripper, (ii) a functional basis $\mathcal{B}_f$ to act on the target object, and (iii) a trajectory $\mathcal{Q}$ directing the functional basis to move towards the target object. 
Given a tool partitioned into a set of sub-meshes $\{\mathcal{M}_i\}$, a sampling process assigns one sub-mesh as $\mathcal{B}_a$ and another as $\mathcal{B}_f$, as illustrated in \cref{fig:tool_basis}. The surface normal vector $\boldsymbol{n}$ at the center of the corresponding sub-mesh indicates the direction for the robot's gripper to approach or for the tool to act on the target object. 
Assuming the robot can firmly grasp the tool at $\mathcal{B}_a$, generating a tool-use strategy $\mathcal{S}$ can be formulated as a motion planning problem that finds a collision-free trajectory $\mathcal{Q}= \boldsymbol{q}_{1:T}$ given $\mathcal{B}_a$ and $\mathcal{B}_f$.

\subsection{\texorpdfstring{\ac{vkc}}{} for Motion Planning}\label{sec:fd_planning}

The theory of body schema~\cite{gallagher2006body} suggests that humans can extend the body's representation to incorporate an external object and treat it as part of their limb for efficient motions and manipulations, which plays a significant role in tool-use~\cite{holmes2006beyond}. This idea has been introduced to the robotics community to represent robot structures and guide robot's behaviors~\cite{hoffmann2010body}. Recent modeling approaches adopting \ac{vkc}~\cite{jiao2021efficient,jiao2021consolidated} provide an effective means to model robot tool-uses: By inserting a virtual joint between robot end-effector and tool's $\mathcal{B}_a$, the kinematics and dynamics of the robot and the tool are integrated, and their motions are planned collectively, resulting in more coordinated motion and higher planning success rate~\cite{jiao2021efficient,jiao2021consolidated}.

We first adopt an articulated body algorithm~\cite{featherstone2014rigid} to compute the forward dynamics analytically for the constructed \ac{vkc}. Next, the objective of the motion planning for robot tool-use is formulated by optimal control:
\begin{align}
    \small
    \underset{x,u,T}{\min} &\ \int_0^T L(x(t), u(t)) dt + \phi(x(T)) \label{eqn:objective} 
\end{align}
\begin{align}
    \small
    L(x(t), u(t)) &= \dot{q}^\top W_{\dot{q}}\dot{q} + u^\top W_{u}u,\\
    \phi(x(T)) &= T, \quad{}T\in\mathbb{R}^+,
    \label{eqn:time_horizon}
\end{align}
where $W_{\dot{q}}$ and $W_{u}$ are weight matrices for joint velocities and joint torques, $u: \mathbb{R}\rightarrow\mathbb{R}^{n}$ the control input consisted of joint torques, $\phi(x(T))$ measures the quality of the terminal state, particularly, we penalize the total elapsed time $T$. $x: \mathbb{R}\rightarrow\mathbb{R}^{2n+2m+1}$ is the state variable, which includes (i) joint positions $q$ and velocities $\dot{q}$ of a manipulator with $n$ \ac{dof}, (ii) $q$ and $\dot{q}$ of underactuated joints in a tool, and (iii) the virtual joint at the grasp point with a total of $m$ \ac{dof}. \cref{eqn:objective} penalizes the weighted quadratic cost on joint velocity and torques for the entire trajectory and the total elapsed time.

During the motion planning, we further impose several safety constraints:
\begin{align}
    \small
    &\dot{x}(t) = f(x(t), u(t)),\quad{}t\in[0,T] \label{eqn:fwd_dyn} \\
    &g(x(t),u(t)) = 0,\quad{}t\in[0,T]
    \label{eqn:state} \\
    &x_{lb} \leq x(t) \leq x_{ub},\quad{}t\in[0,T] \label{eqn:joint_limit} \\
    &u_{lb} \leq u(t) \leq u_{ub},\quad{}t\in[0,T] \label{eqn:ctrl_limit} 
\end{align}
where \cref{eqn:fwd_dyn} is the system dynamics, \cref{eqn:state} is a task-dependent constraint for tool-use, \cref{eqn:joint_limit} and \cref{eqn:ctrl_limit} are safety constraints that bound the robot workspace and control limit.

\subsection{Goal Specification}\label{sec:goal_spec}

Formally, the goal for a tool-use is expressed as:
\begin{equation}
    f_\text{task}(n_\text{T}(\mathcal{G}), \text{VKC}) \Rightarrow g(\cdot),
    \label{eqn:task}
\end{equation}
where $n_\text{T}(\mathcal{G})$ is a set of physical properties that are essential to the task, to be detailed in \cref{sec:sim_learn}. $f_\text{task}$ maps these physical properties and \acp{vkc} (as constructed in \cref{fig:robot_tool_vkc}) to a constraint function $g$ for motion planning. The intuition is for the robot to emulate those essential physical properties in execution while considering the robot and tool's kinematics and dynamics. 

To be more specific, let us take the walnut cracking task as an example. Given the goal position where the contact occurs $p_{g}$, the tool should act on the target object with a velocity vector $\mathbf{v}_\text{tool}$ and the tool's orientation $\mathbf{d}_\text{tool}$ (to be detailed in \cref{sec:goal}), both represented in world frame. \cref{eqn:new_goal} first finds a possible robot goal pose $q_g$ through solving inverse kinematics to regulate the tool's orientation when contacting the target object:
\begin{align}
    \frac{f_z(q_g)\cdot \mathbf{v}_\text{tool}}{||f_z(q_g)||\cdot||\mathbf{v}_\text{tool}||} = \text{cos}(\mathbf{d}_\text{tool}),
    \label{eqn:new_goal}
\end{align}
where $f_z:\mathbb{R}^n\rightarrow\mathbb{R}^3$ finds the surface normal of $\mathcal{B}_f$. Then, the goal joint velocities are computed by:
\begin{align}
    \dot{q}_{g} &= J_\text{VKC}^\top(J_\text{VKC}J_\text{VKC}^\top)^{-1}\mathbf{v}_\text{tool},
    \label{eqn:goal}
\end{align}
where $J_\text{VKC}$ is the geometric Jacobian from the robot's base frame to the tool's functional basis at the joint position $q_g$. Finally, \cref{eqn:task} can be expressed in terms of joint velocity \wrt two constraint functions $q_g$ and $\dot{q}_g$:
\begin{align}
    g_q(x(t),u(t))=& x_q(T) - q_g = 0 \\
    g_{\dot{q}}(x(t),u(t))=& x_{\dot{q}}(T) - \dot{q}_g = 0
\end{align}

\section{Simulation and Learning}\label{sec:sim_learn}

This section starts with the technical background of physics-based simulation, followed by how it reproduces fine-grained physical properties and helps understand tool-uses events.

\subsection{Background}

Solid simulation approximates objects' physical status. It is oftentimes formulated with the \acf{fem}~\cite{zienkiewicz2000finite}, which discretizes each object into small elements with a discrete set of sample points as the degree-of-freedoms. Mass and momentum conservation equations are discretized on the mesh and integrated overtime to capture the dynamics. This paper utilizes an \ac{ipc} handling method~\cite{li2019decomposed,li2020incremental,li2020codimensional}, a state-of-the-art \ac{fem}-based simulator, to address the difficulty of simulating non-smooth contacts between a tool and a target object. To further support object fracture during tool-uses, our simulator measures the displacement of every pair of points that both connect to all the nodes of a triangle on the mesh. If the displacement relative to their original distance exceeds a certain strain threshold, we mark the triangle in-between as separated.

\begin{figure}[t!]
    \centering
    \begin{subfigure}[b]{\linewidth}
        \centering  
        \includegraphics[width=\linewidth]{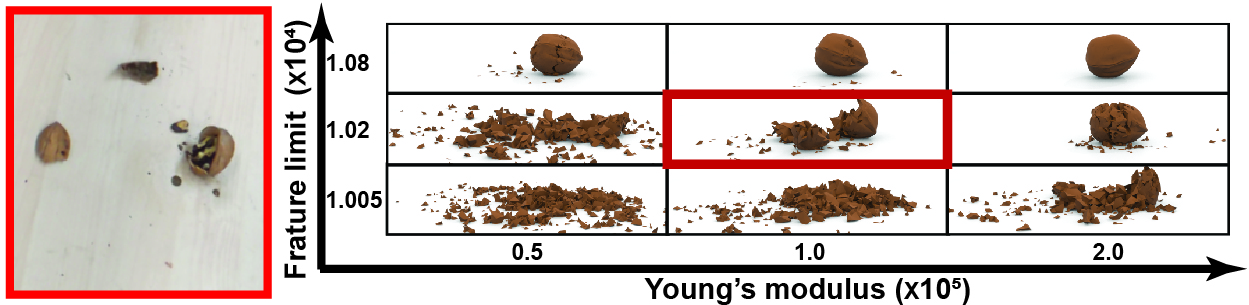}
        \caption{Comparisons between the effect produced by experiments (left) and simulations (right) with different fracturing limits and Young's modulus.}
        \label{fig:sim_effect}
    \end{subfigure}%
    \\%
    \begin{subfigure}[b]{\linewidth}
        \centering  
        \includegraphics[width=\linewidth]{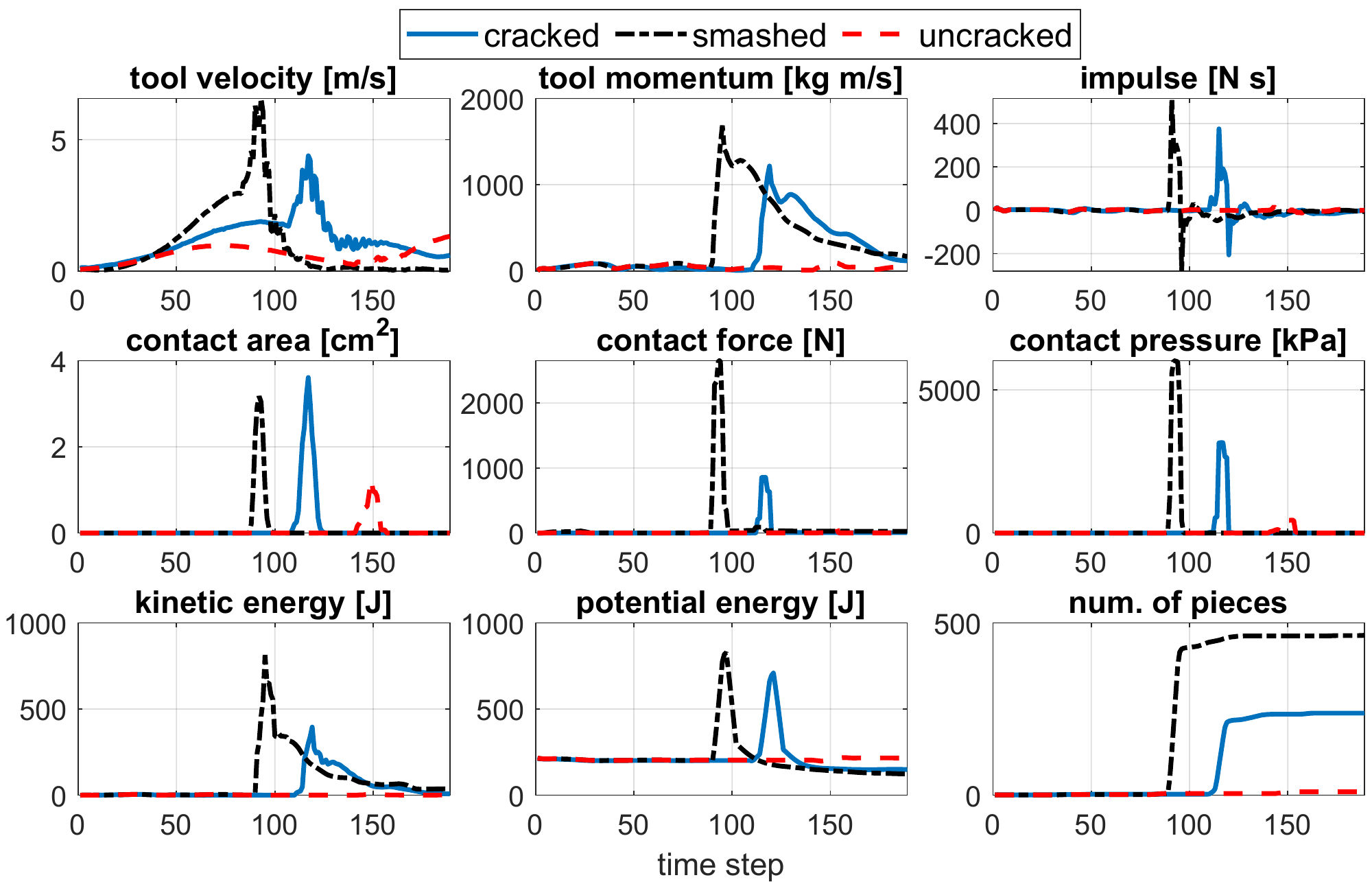}
        \caption{Fine-grained physical properties evolved in time reflect the different physical effects among uncracked, cracked, and smashed.}
        \label{fig:sim_physics}
    \end{subfigure}%
    \caption{\textbf{Examples of qualitative and quantitative results produced by the \ac{fem}-based simulation.} (a) We qualitatively choose the parameters (in red) that best match the final effect of observed tool-use events. (b) By adopting an \ac{fem}-based simulator, the data collection process records physical properties evolved in time.}
\end{figure}

\subsection{Reproducing Effect}\label{sec:data_collect}

To produce similar effects in the simulation that sufficiently match those in the physical world, some parameters governing an object's material property need to be appropriately set. In particular, Young's modulus reflects the object's stiffness\textemdash{}the stiffer the object is, the harder for it to deform or fracture, and fracture limit determines the number and the size of segments a large piece will fracture into. \cref{fig:sim_effect} qualitatively shows how the resulting effects vary given different Young's modulus and fracture limit. These two parameters are calibrated such that the simulated effects match the observation in physical world.

We use two VIVE Trackers to capture the tool-use events. One to track the movement of the tool (\eg, a hammer), and another placed on the table serving as the reference point for the target object (\eg, a walnut). Both VIVE trackers are calibrated such that their relative poses and captured trajectories are expressed in the same coordinate frame, with a time step of the inverse of their sampling frequency. The meshes of the target object and the tool are pre-scanned using an RGB-D camera. Combining scanned meshes and captured trajectories, we can fully reconstruct an observed event in both space and time and further simulate the effects of the target object both visually and physically. Examples of keyframes of the collected data with corresponding simulated results are shown in \cref{fig:motivation}a. \cref{fig:sim_physics} visualizes the continuous numerical values of some notable physical properties obtained from simulation. Of note, capturing how the object's status changes and its physical properties evolve over time is highly challenging, if not impossible, using visual information alone. 

\subsection{Learning Essential Physical Properties}

We quantize the space of physical properties into three different levels; see \cref{fig:motivation}a for an illustration. (i) \textbf{Action} (in blue) includes the trajectory data (position and orientation) directly observed in tool-use events and its velocity and acceleration calculated by finite-difference; these properties are usually controllable by robots. (ii) \textbf{Simulation} (in green) includes the physical properties estimated by the simulation given the observed Action. (iii) \textbf{Effect} (in red) includes the physical properties representing the tool-use effect. In the case of cracking and cutting tasks, we represent the effect by the number of pieces the target object transforms into.

Given various physical properties estimated and reproduced by the simulation, a robot has to learn how much these properties contribute to the success of the task and distill knowledge at all three levels, such that it can plan its motion in new and even unseen scenarios. To encode the connections across all three levels of physical properties, we propose to learn a \textbf{\ac{prg}} representation through symbolic regression~\cite{schmidt2009distilling,udrescu2020ai}. Specifically, setting the Effect as the target variable $y$, the symbolic regression is tasked to find a valid expression of $y$ using the set of given variables $\textbf{x}$ in Simulation and Action: $y=f(\textbf{x})$. To prevent over-fitting, we further balance the expression's complexity (\ie, how many physical properties are involved) and accuracy (\ie, how well it expresses the target variable). As such, the relations in \ac{prg} is sparse and only involve a small subset of the variables that succinctly express the target variable.

\begin{figure}[t!]
    \centering
    \includegraphics[width=\linewidth]{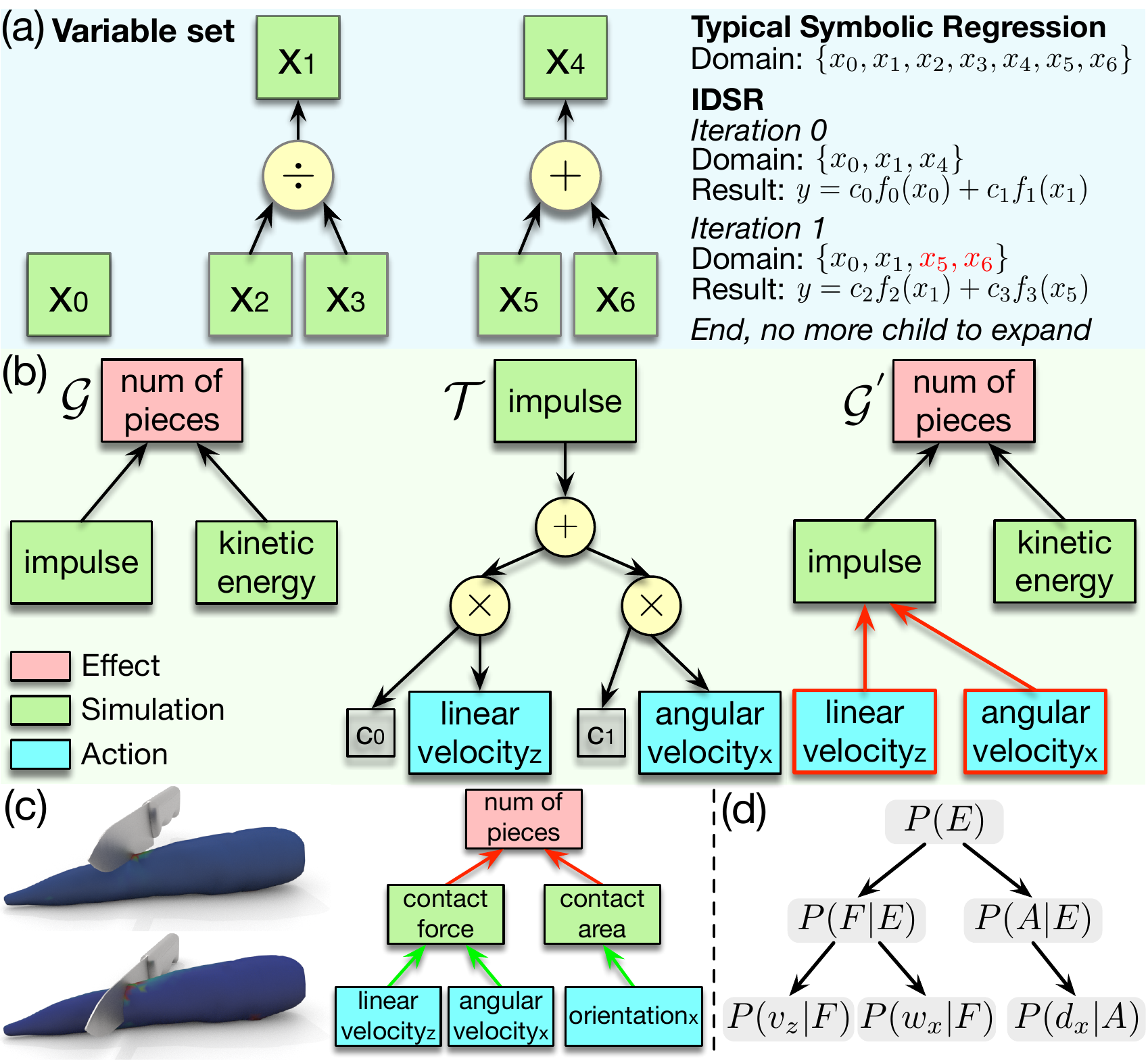}
    \caption{\textbf{Learning relations among physical properties using \ac{idsr}}. (a) An example of deepening the variable domain. Since $x_4$ is not included in the resulted expression in the iteration 0, it is thus removed, and its children are added to the domain in the next iteration. (b) An example of constructing \ac{prg}. $\mathcal{G}^{'}$ isthe updated graph after inserting the expression $\mathcal{T}$ into the previous graph $\mathcal{G}$; newly added nodes and edges are highlighted in red. (c) The \ac{prg} constructed for the cutting task. (d) Inferring necessary values at the Action level for the goal specification in planning.}
    \label{fig:concept_graph}
\end{figure}

Typical symbolic regression problems oftentimes have a large search space. To tackle it, we devise an \ac{idsr} algorithm, a variant of symbolic regression, that utilizes the hierarchical information among physical properties at each level to prune the searching space. Specifically, as illustrated in \cref{fig:concept_graph}a, typical symbolic regression algorithms directly explore the entire domain with all variables, whereas the proposed \ac{idsr} would iteratively deepen the domain based on the hierarchy among them. If one variable is not selected after an iteration, the domain will replace it with its child variables and reiterate the algorithm, and the resulting expression will only be updated if those child variables play a more significant role. This process continues until all the variables from one level in the domain are selected, or non-selected variables have no child. \cref{alg:idsr} outlines the procedure.

In the case of cracking a walnut (see \cref{fig:concept_graph}b), only after the set of relations between \textit{Effect} and \textit{Simulation} is explored would the algorithm subsequently identify the set of relations between \textit{Simulation} and \textit{Action}, expanding the \ac{prg}. As a result, this algorithm design saves the memory compared to conventional symbolic regression algorithms while preserving the full capability of distilling the essential relations among variables. The sub-graph highlighted in red in \cref{fig:motivation}a shows the learned \ac{prg} of cracking a walnut, wherein the edge thicknesses are proportional to the physical properties' contribution to the effect. In another task of cutting a carrot by half using a knife (see \cref{fig:concept_graph}c), the \ac{idsr} algorithm identifies the \textit{contact area} governed by the \textit{orientation} as an essential physical property, since the deviation from a proper orientation range may lead to the increment of contact area.

\begin{algorithm}[b!]
    \small
    \caption{\ac{idsr}}
    \label{alg:idsr}
    \KwData{
        Data samples: $\mathcal{D}$. Target variable: $v_{t}$. 
        Variable set: $\mathcal{V}$ 
    }
    \KwResult{
        Best matched expression tree: $\mathcal{T}$
    }
    \BlankLine
    $Domain \gets \{\texttt{AllRoots($\mathcal{V}$)}\}$ 
    \While{not terminate}{
        $terminate \gets \texttt{True}$\;
        \Comment{Symbolic regression on Domain}
        $\mathcal{T} \gets \texttt{SR($\mathcal{D},v_t, Domain$)}$\;
        $diff \gets Domain~\backslash~\mathcal{T}.\texttt{leaveSymbols()}$\;
        \Comment{Deepening the searching domain}
        \ForEach{v in diff}{
            \If{v has child}{
                $Domain.\texttt{add($v.\texttt{children()}$)}$\;
                $Domain.\texttt{remove($v$)}$\;
                $terminate \gets \texttt{False}$\;
            }
        }
    }
    \Return $\mathcal{T}$ \Comment{Return the latest $\mathcal{T}$}
\end{algorithm}

\subsection{Reasoning about Goal Specification}\label{sec:goal}
The $\mathcal{G}$ identified by \ac{idsr} is still insufficient to support the proposed planning scheme because it only deduces the relation among those physical properties in a symbolic level, \ie \textit{velocity} for the task of cracking a walnut, and both \textit{velocity}  and \textit{orientation} $\mathbf{d}_\text{tool}$ for cutting as shown in \cref{fig:concept_graph}bc. The corresponding values of $\mathbf{v}_\text{tool}$ and/or $\mathbf{d}_\text{tool}$ applicable for robot planning is not determined yet.

To address this issue, we devise a sequential inference pipeline based on learned $\mathcal{G}$. As illustrated in \cref{fig:concept_graph}d, by modeling the values of observed effect as a Gaussian distribution $P(E)$, a \ac{gmm} is learned to capture the joint probability between the effect and an identified physical property according to $\mathcal{G}$, \eg $P(E,F)$ for effect and contact force in \cref{fig:concept_graph}cd, using the EM algorithm~\cite{dempster1977maximum}. Specifically, the mixture models are fitted on the data obtained from the simulator that reproduces human demonstrations. Next, given a desired effect, inferring specific values of \textit{contact force} is performed by drawing samples from the distribution $P(F|E) = P(E,F) / P(E)$~\cite{bishop2006pattern}, and a velocity in $z$ direction $v_z$ is subsequently obtained by sampling from $P(v_z|F)$ following the same protocol. Eventually, this process produces the necessary values at the Action level ($\mathbf{v}_\text{tool}$ and $\mathbf{d}_\text{tool}$) as goal specifications for \cref{eqn:new_goal,eqn:goal}.

\section{Experiments}\label{sec:exp}

\begin{figure}[t!]
    \centering
    \includegraphics[width=\linewidth]{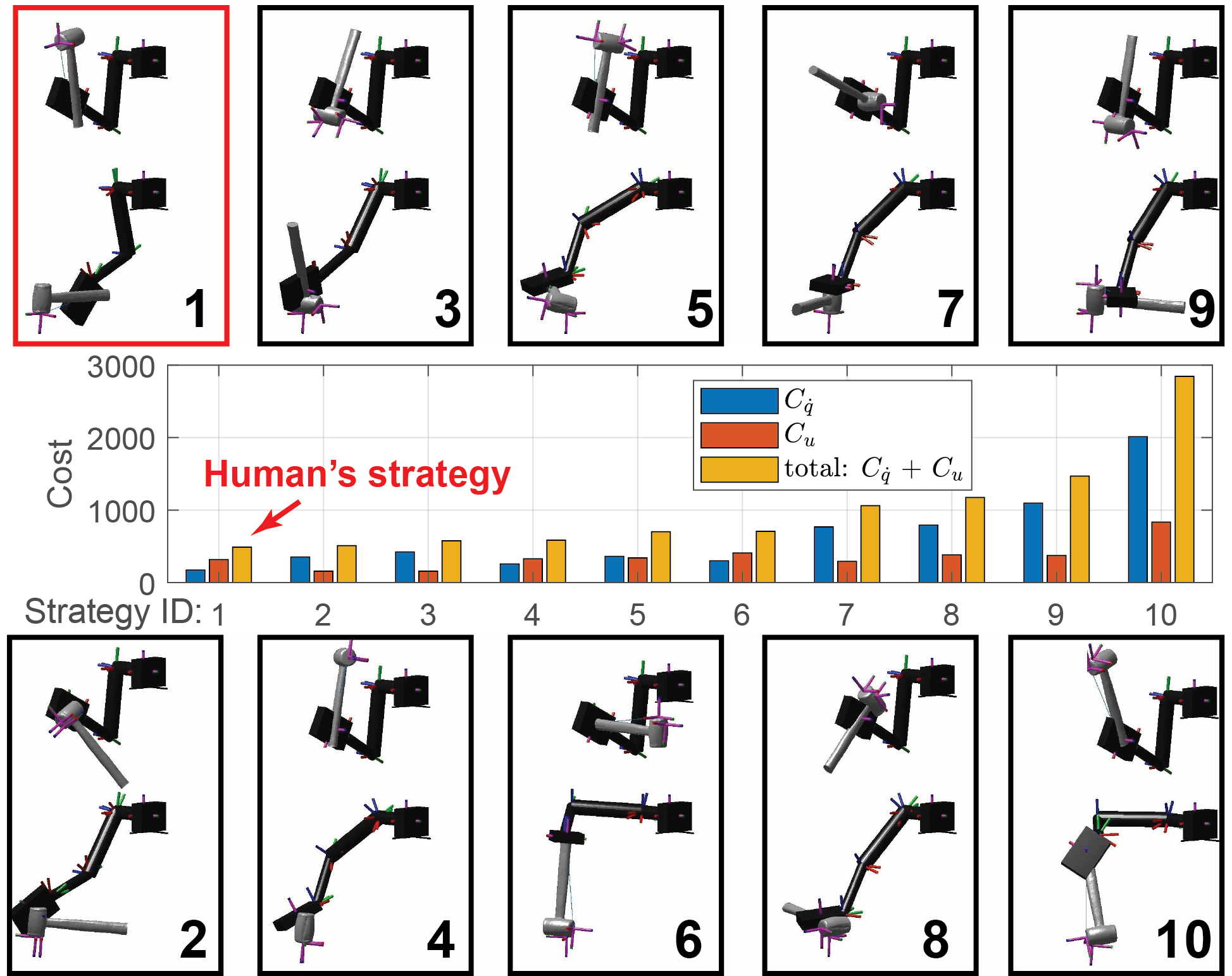}
    \caption{\textbf{Different strategies of tool-use using an approximated human arm model}. $\mathcal{B}_a$s and $\mathcal{B}_f$s are sampled from partitioned regions on the hammer, and the trajectory $\mathcal{Q}$ is produced by the optimal control using \acp{vkc}. The optimal strategy (in red) indeed follows human intuition of operating a hammer. $C_{\dot{q}}$ is the trajectory smoothness cost, and $C_u$ is the joint torque effort cost.}
    \label{fig:exp1_human}
\end{figure}

We conduct three sets of studies regarding different types of manipulators with various settings. Using a human arm model~\cite{wu2005isb}, we first validate that our planning scheme produces a feasible tool-use strategy identical to human choices; see \cref{sec:exp1}. Next, we show that our proposed framework produces diverse tool-use strategies for Baxter arm and UR5 manipulator under different scenarios; the most effective ones in terms of least joint effort are demonstrated in \cref{sec:exp2}. Finally, the produced strategies are fed to simulations for robot planning and execution; see \cref{sec:exp3}. Experimental results verify that the framework indeed captures the essential physical properties, capable of converting these learned relations into goal specification, resulting in the success of motion planning and task completion.

In all experiments, we solve the motion planning problem defined in \cref{sec:definition} by CasADi~\cite{andersson2019casadi} with the OpenOCL~\cite{koenemann2017openocl} support. A tool-use is considered invalid if the planner cannot produce a feasible solution. We assume the manipulators' bases are fixed. The tool structures are scanned by an RGB-D camera and reconstructed into watertight meshes, and the tool's material is homogeneous. For fair comparisons, the target object (\eg, walnut) is placed at the same location within the operational space for each type of manipulator, and the initial pose of the manipulator is identical across all trials. 
In each trial, the target object has 1229 mesh vertices and is simulated for 200 time steps. The simulation runs on a 16-core AMD Ryzen 9 5950X machine and the average run time for one trial is 77.08 minutes with parallelization for the linear system computations~\cite{li2020incremental}. Algorithm-wise parallelization for \ac{fem} still remains an open problem.

\begin{figure*}[t!]
    \centering  
    \begin{subfigure}[b]{\linewidth}
        \centering  
        \includegraphics[width=\linewidth]{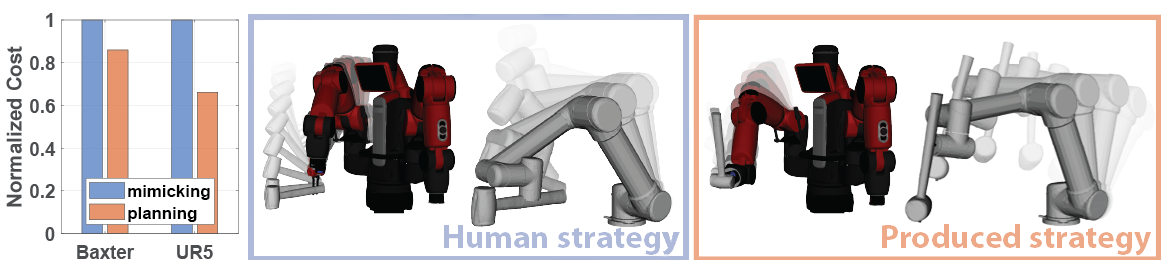}
        \caption{Comparison of joint effort costs in hammering between mimicking human's strategy and the most effective one produced by our framework.}
    \end{subfigure}%
    \\%
    \begin{subfigure}[b]{\linewidth}
        \centering  
        \includegraphics[width=\linewidth]{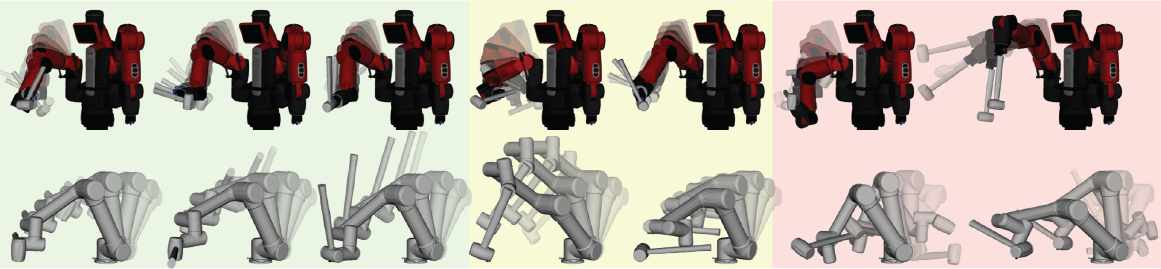}
        \caption{Examples of various strategies to use the hammer.}
    \end{subfigure}%
    \caption{\textbf{Generated various strategies in using a hammer}. (a) Given an inferred velocity vector acting on the walnut, the best tool-use strategy for each robot found by our framework is more efficient than simply mimicking human's strategy, indicated by lower cost. (b) Other strategies found by the proposed framework: low cost (in green), high cost (in yellow), and invalid with violation of constraints (in red).}
    \label{fig:exp2_robot}
\end{figure*}

\subsection{Validating Optimal Planning by Task Efficiency}\label{sec:exp1}

In this experiment, we evaluate whether the optimal control-based planning scheme is effective by comparing the produced tool-use strategies with that of human's rational choice, which should be regarded as near-optimal. Using the human arm model~\cite{wu2005isb} that consists of 7 \acp{dof} (3 for shoulder, 2 for elbow, and 2 for wrist) with corresponding arm's physical properties (\ie, mass, inertia) measured by human subjects, we sample various combinations of $\mathcal{B}_a$ and $\mathcal{B}_f$ and produce the corresponding tool-use trajectories. \cref{fig:exp1_human} shows initial and final arm postures and their computed costs of replicated human tool-uses and nine examples of alternative solutions.

Our results show that Strategy 1 is the most efficient one. Compared with a conventional swinging action, holding hammerhead reduces the inertia compensated by actions, resulting in a lower joint torque effort costs $C_u$s in Strategy 2 and 3. However, the trajectory smoothness costs $C_{\dot{q}}$s are higher as a larger acceleration is required to reach the goal velocity, making their total costs higher than the cost in Strategy 1. Since both Strategy 4 and 5 start from a similar $\mathcal{B}_a$ as in Strategy 1 followed by a swinging trajectory, their $C_u$s are similar to that of Strategy 1; however, their $C_{\dot{q}}$s are higher since their $\mathcal{B}_f$s do not well aligned with arm postures. Strategy 6 to 10 are some less typical examples with high $C_u$s and $C_{\dot{q}}$s; we seldom observe these strategies in real life. Together, these results indicate that our planning scheme can produce an efficient tool-use trajectory with underlying rationales akin to human tool-use behaviors, and thus we expect it to uncover similar insights into robot tool-uses.

\subsection{Effective Tool-Uses}\label{sec:exp2}

After validating our optimal control-based planning scheme, we test the efficacy of tool-use strategies derived from learned physical properties using two different robots (a Baxter robot and a UR5 manipulator) in two tasks (cracking and cutting).

Due to significant differences in kinematic structures, the observed human strategy of tool-uses may not be ideal for robots. In \cref{fig:exp2_robot}a, two robots first mimic human's strategy. Specifically, the robots select the observed human's $\mathcal {B}_a$ and $\mathcal{B}_f$ and mimic the observed trajectory $\mathcal{Q}$ by inverse kinematics to operate the hammer. The resulting costs are higher than those of the best strategy found by our framework; the ones found by the proposed framework are dramatically different but more effective for the robots. \cref{fig:exp2_robot}b further displays some other tool-use strategies with low-cost (effective), high-cost (ineffective), or are invalid by violating constraints.

\begin{figure*}[t!]
    \centering
    \begin{subfigure}[b]{0.5\linewidth}
        \centering  
        \includegraphics[width=\linewidth]{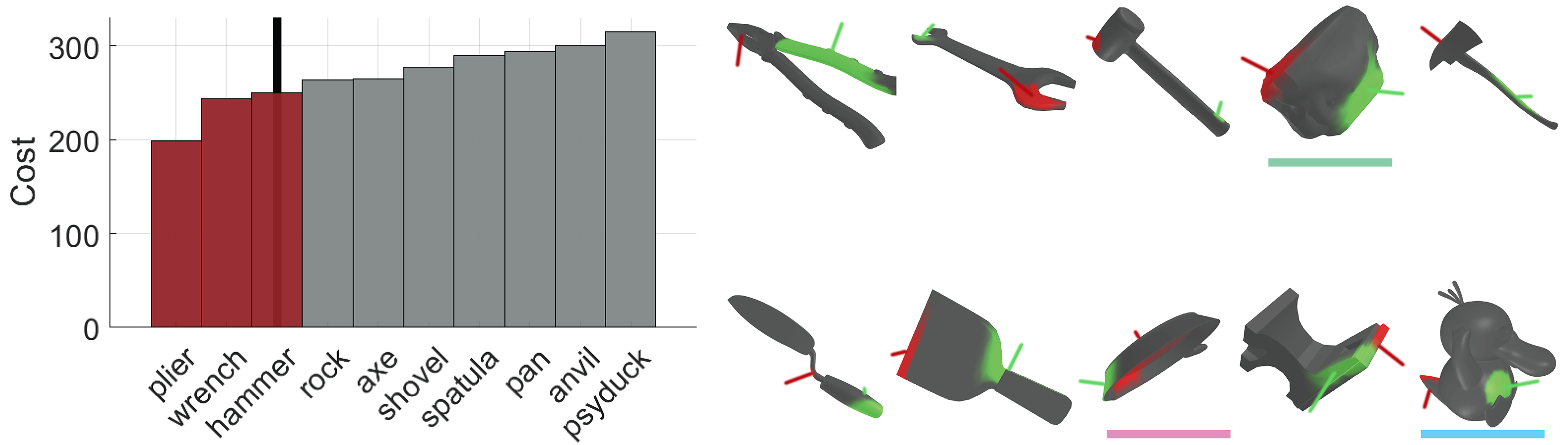}
        \caption{Best strategies found for the Baxter robot.}
        \label{fig:various_tool_baxter}
    \end{subfigure}%
    \begin{subfigure}[b]{0.5\linewidth}
        \centering  
        \includegraphics[width=\linewidth]{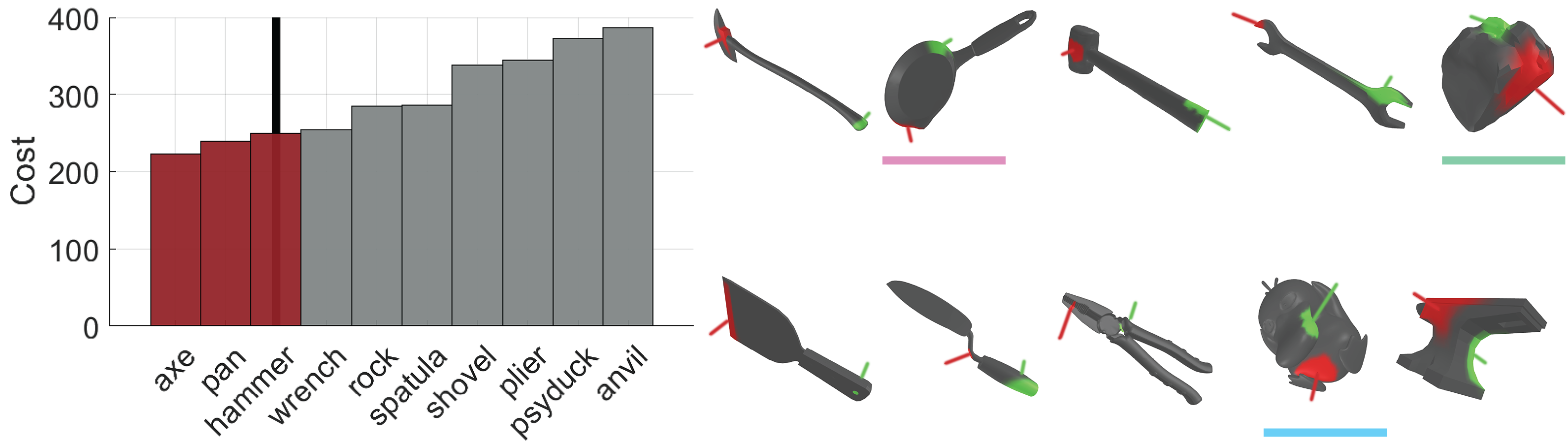}
        \caption{Best strategies found for the UR5 manipulator.}
        \label{fig:various_tool_ur5}
    \end{subfigure}%
    \\
    \begin{subfigure}[b]{\linewidth}
        \centering  
        \includegraphics[width=0.95\linewidth]{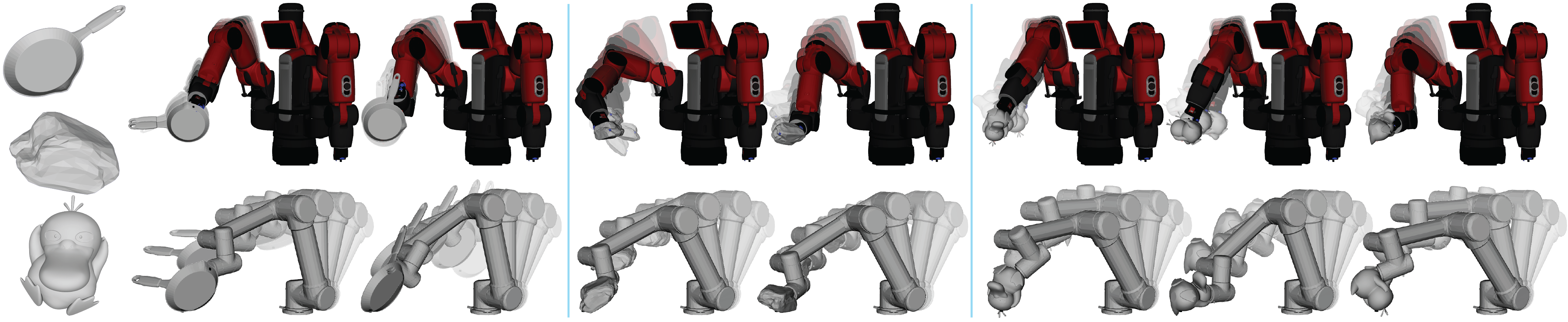}
        \caption{Trajectories of using some of the unconventional tools: a pan, a rock, and a toy.}
        \label{fig:various_tool_new_object}
    \end{subfigure}%
    \caption{\textbf{Effective tool-uses with unseen objects for the walnut-cracking task}. (a)(b) The best strategies (least cost) for ten different objects to crack a walnut use a Baxter robot and a UR5 manipulator, respectively. (c) Examples of valid trajectories of the Baxter robot (upper) and a UR5 manipulator (lower) using a pan (left), a piece of rock (middle), and a Psyduck toy (right).}
    \vspace{-3pt}
    \label{fig:exp4}
\end{figure*}

Our framework is generic and generalizable to more challenging cases. It can further generate effective strategies using unconventional daily objects. The costs of operating those objects are ranked from low to high in \cref{fig:exp4}a (Baxter) and \cref{fig:exp4}b (UR5). The experiment reveals some objects (piler and wrench for Baxter, and axe and pan for UR5) are surprisingly more handy for robots compared with the hammer (indicated by the black bar). We further visualize the executed trajectories in \cref{fig:exp4}c. Of note, the same pan is more suitable for UR5 as the cost of operating it is lower than using a hammer but not that effective for Baxter. In comparison, the efficiency of using the rock and the toy (Psyduck) are similar for both robots. These results demonstrate that our learning and planning framework enables a situational tool-use skill for various robots.

\begin{figure}[t!]
    \begin{subfigure}[b]{0.333\linewidth}
        \centering
        \includegraphics[width=\linewidth]{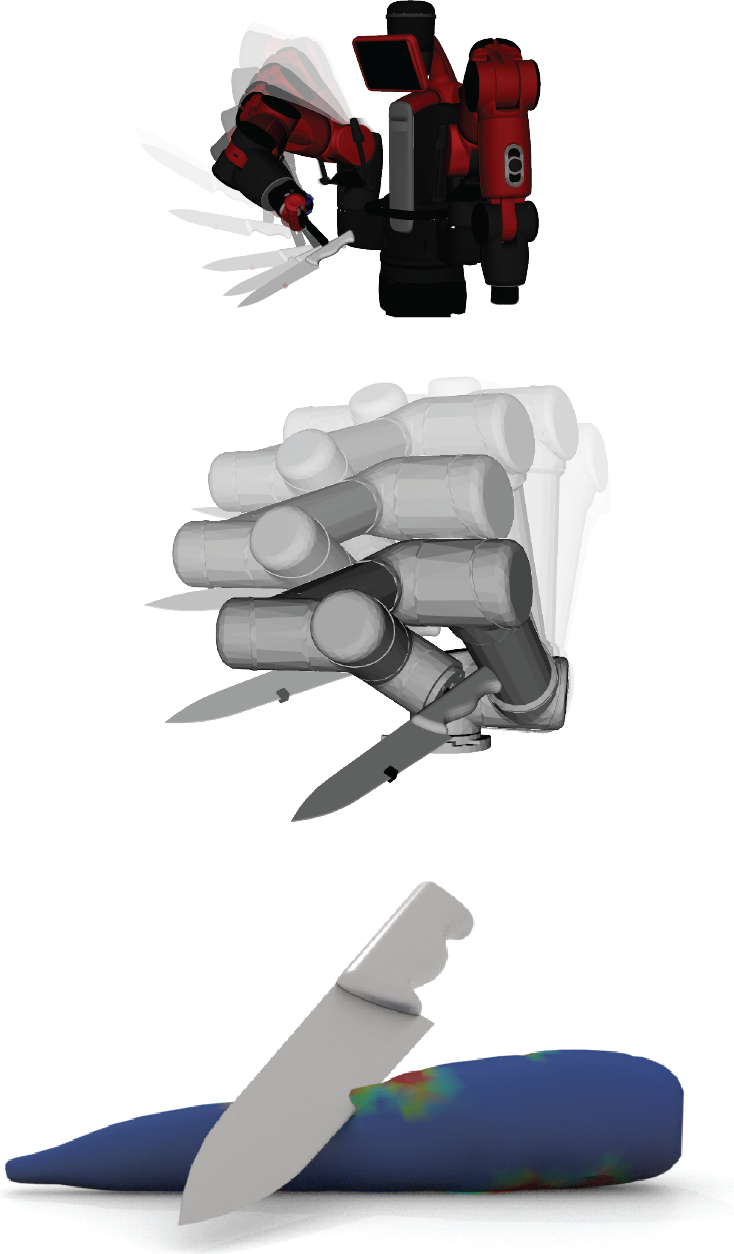}
        \caption{}
    \end{subfigure}%
    \begin{subfigure}[b]{0.333\linewidth}
        \centering  
        \includegraphics[width=\linewidth]{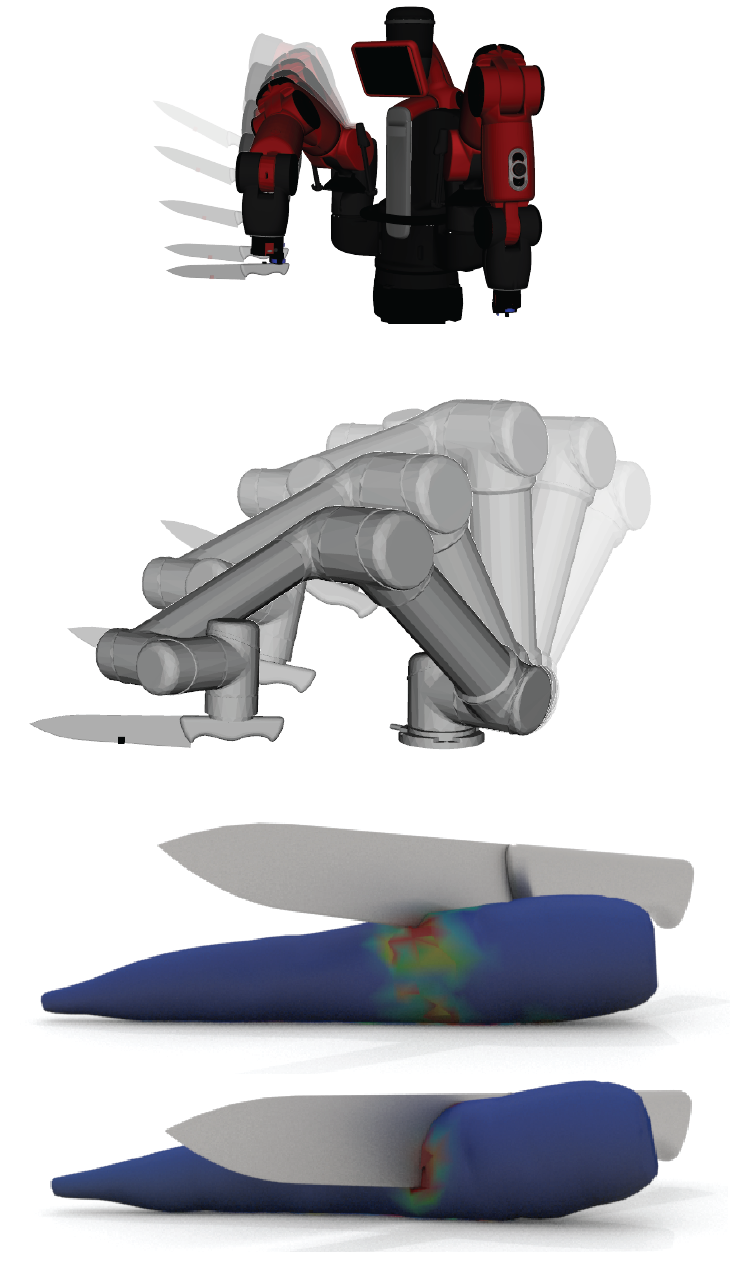}
        \caption{}
    \end{subfigure}%
    \begin{subfigure}[b]{0.333\linewidth}
        \centering  
        \includegraphics[width=\linewidth]{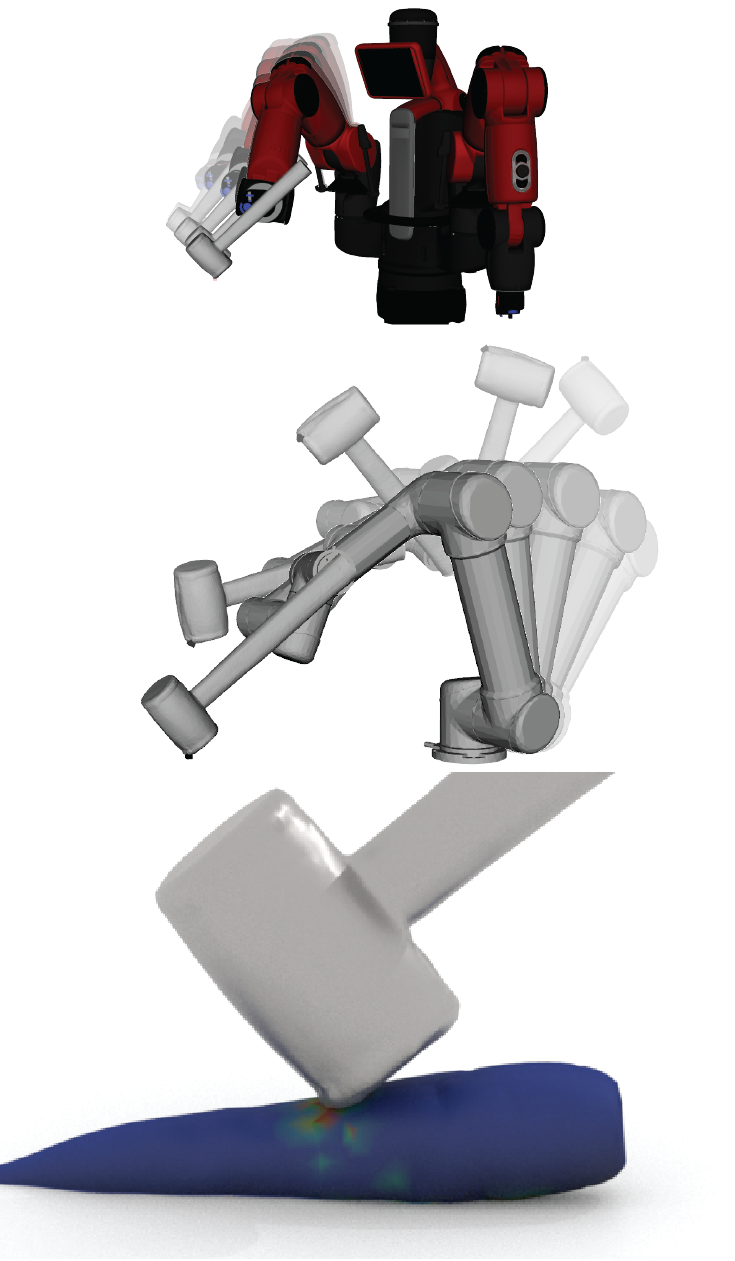}
        \caption{}
    \end{subfigure}%
    \caption{\textbf{Tool-use strategies for cutting the carrot}. (a) Robots fail to accomplish the task without incorporating a tool's orientation. (b) The successful use of a knife requires incorporating orientation properties as learned in \cref{fig:concept_graph}c. (c) Even using an object (a hammer) unsuitable for this task, our framework still produces an effective strategy by finding a tool orientation that minimize contact.}
    \vspace{-3pt}
    \label{fig:exp3_cut}
\end{figure}

In another task of cutting carrot, both robots do not perform well if concerned only about the \textit{velocity} as they did in walnut-cracking; the target object will not align with the knife's blade properly as illustrated in \cref{fig:exp3_cut}a. By incorporating tool's \textit{orientation} as uncovered in \cref{fig:concept_graph}c, the robots overcome this deficiency and produce desired effects successfully; see \cref{fig:exp3_cut}b. Compared with the walnut-cracking task, the cutting task poses greater challenges in selecting unseen objects as tools since not all objects can lead to task completion; \ie, one cannot use a hammer to successfully cut a carrot as a knife does. Yet in \cref{fig:exp3_cut}c, the result still demonstrates the robots' reasonable efforts in this difficult situation by choosing a sharp edge to contact with the object, showing that our framework successfully captures the essential physics in tool-uses and leverages them in producing its own strategies.

\subsection{Testing Robot Tool-use in Simulation}\label{sec:exp3}

Finally, we evaluate how well the best strategy found by the proposed framework (\eg, produced strategies in \cref{fig:exp2_robot}a) can be executed in the simulator. This step is crucial as it separates the proposed framework from purely vision-based methods.

Since no existed work can solve the proposed task, we design a kinematic-based motion planner as a baseline that accounts for the physical properties involved in the task. In the case of the walnut-cracking task, the baseline needs to plan a trajectory that moves the functional basis of the tool to the center of the walnut while keeping its surface normal aligned with the gravity direction. Fifty trials are simulated for both the Baxter robot and the UR5 manipulator using trajectories produced by the baseline and the proposed framework, and the parameters governing walnut's fracturing properties in each trial are set based on the values shown in \cref{fig:sim_effect} with a randomness of $10\%$ for variations.

\begin{figure}[t!]
    \centering
    \includegraphics[width=\linewidth]{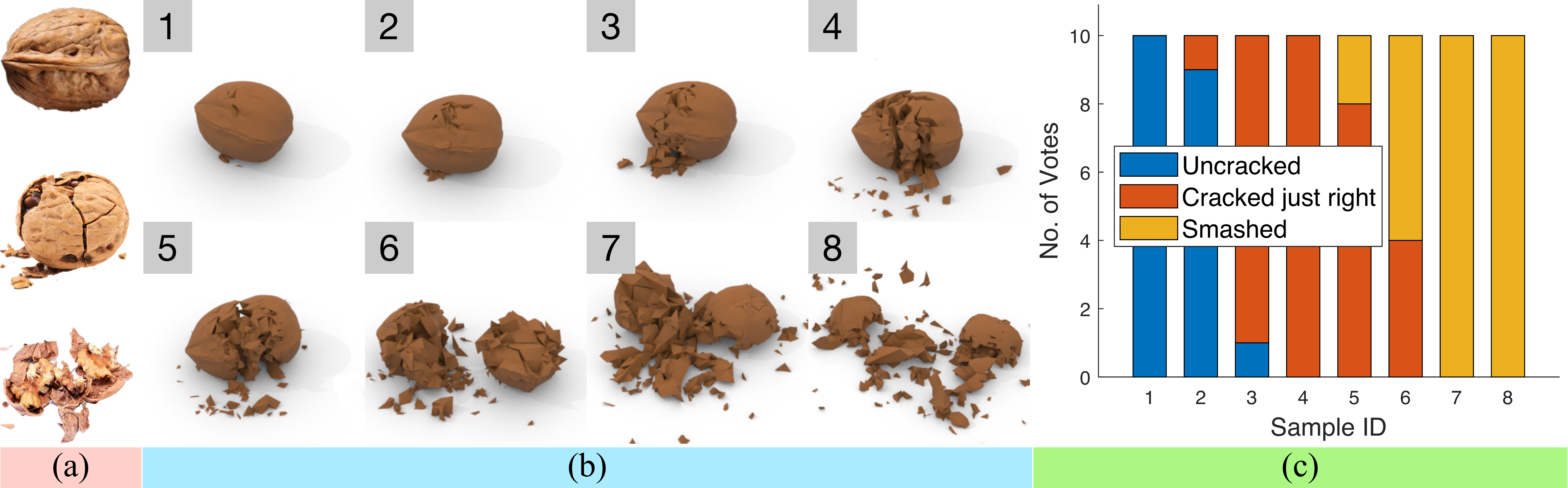}
    \caption{\textbf{Human evaluation of classifying the status of simulated execution results}. 
    (a) After presenting three instances of walnut being uncracked, cracked just right, and smashed, (b) participants are asked to classify observed simulation results (eight samples for illustration) into one of these three statuses. (c) Sample 3 to 5 are considered successful as most participants regard them as cracked.}
    \vspace{-3pt}
    \label{fig:human_votes}
\end{figure}

Due to the lack of quantitative evaluation of the performance of the walnut cracking task, we conducted a human study to compare the results between the baseline and the proposed framework. Ten participants were recruited online and asked to classify the total of $200$ simulated execution results into one of the three statuses based on three instances shown in \cref{fig:human_votes}a. An execution is considered successful if more participants regard the walnut's status as cracked. \cref{fig:human_votes}bc show eight examples of the results based on the human study. The success rates are shown in \cref{tab:success_rate}, demonstrating the necessity of understanding the physics in tool-use. Together, the results show the proposed framework indeed enables a better understanding of complex physical events that occurred during the tool-uses and successful productions of tool-use behaviors for robots.

\begin{table}[b!]
    \small
    \centering
    \caption{Success Rate in Cracking Walnut in Simulator}
    \begin{tabular}{c c c}
        \toprule
        \textbf{Robot Type} & \textbf{Baseline}  &  \textbf{Proposed}
        \\ \midrule
        Baxter & $14\%$ & $62\%$\\
        UR5 & $16\%$ & $52\%$\\
        \bottomrule
    \end{tabular}
    \label{tab:success_rate}
\end{table}

\section{Conclusion and Discussion}\label{sec:conclusion}

We presented a learning and planning framework for robots to understand the physics behind tool-use events and generate tool-use strategies suitable for the robots' own kinematics and dynamics. A physics-based \ac{fem} simulator was developed to generate physical properties in a continuous manner, from which we devised an \ac{idsr} algorithm to learn the essential properties critical to the success of the task. By formulating the learned properties into an optimal control-based motion planning scheme, our experiments demonstrated that the proposed framework allows robots to find tool-use strategies different from human demonstrations when handling seen and unseen objects, with better efficiency measured by least joint efforts.

While our work is conducted in simulation, our planning scheme outputs torque commands that are possible for deployment on physical robots in the future. As grasping remains an unsolved problem, we plan to incorporate more sophisticated methods (\eg, \cite{liu2022synthesizing}) to generate firm grasp configurations on the tool, such that we can produce more realistic and adaptive tool-uses. The reality gap is another major challenge to realize the physical deployment of the proposed framework. Physics-based simulation is difficult to tune or match the real world precisely. However, it is still a powerful tool for robot understanding and uncovering the task goal.

\textbf{Acknowledgement:} The authors would like to thank Dr. Minchen Li and Dr. Chenfanfu Jiang from the UCLA Mathematics Department for the simulation setup.

{
\bibliographystyle{ieeetr}
\bibliography{reference}
}

\end{document}

%% file: configs.tex
\usepackage{graphicx} \graphicspath{ {figures/} }
\usepackage{amsmath,amssymb,mathabx}
\usepackage{algorithmic}
\usepackage[ruled,linesnumbered]{algorithm2e}
\usepackage{acronym}
\usepackage{enumitem}
\usepackage{booktabs}
\usepackage{hyperref}
\usepackage{bm}
\usepackage{balance}
\usepackage{xspace,setspace}
\usepackage[skip=3pt,font=small]{subcaption}
\usepackage[skip=3pt,font=small]{caption}
\usepackage[dvipsnames]{xcolor}
\usepackage[capitalise]{cleveref}
\usepackage{tabularx,colortbl,multirow,array,makecell}
\usepackage{overpic}
\usepackage{cite}
\usepackage{tikz}

\makeatletter
\DeclareRobustCommand\onedot{\futurelet\@let@token\@onedot}
\def\@onedot{\ifx\@let@token.\else.\null\fi\xspace}
\def\eg{\emph{e.g}\onedot} 
\def\ie{\emph{i.e}\onedot}

\def\wrt{w.r.t\onedot~} 
\def\etal{\emph{et al}\onedot~}
\makeatother

\makeatletter
\def\BState{\State\hskip-\ALG@thistlm}
\makeatother

\makeatletter
\renewcommand{\paragraph}{%
  \@startsection{paragraph}{4}%
  {\z@}{0ex \@plus 0ex \@minus 0ex}{-1em}%
  {\hskip\parindent\normalfont\normalsize\bfseries}%
}
\makeatother

\crefname{algocf}{Alg.}{Algs.}
\Crefname{algocf}{Algorithm}{Algorithms}

\definecolor{gblue}{HTML}{4285F4}
\definecolor{gred}{HTML}{DB4437}

\acrodef{dof}[DoF]{Degree of Freedom}
\acrodef{vkc}[VKC]{Virtual Kinematic Chain}
\acrodef{fem}[FEM]{Finite Element Method}
\acrodef{ipc}[IPC]{Incremental Potential Contact}
\acrodef{prg}[PRG]{Physical Relation Graph}
\acrodef{idsr}[IDSR]{Iterative Deepening Symbolic Regression}
\acrodef{gmm}[GMM]{Gaussian Mixture Model}

\SetKwComment{Comment}{//}{}